# A high-precision underwater object detection based on joint self-supervised deblurring and improved spatial transformer network

Xiuyuan Li※, Fengchao Li, Jiangang Yu, Guowen An
School of Instrument and Electronics, North University of China, Taiyuan, China
E-mail: lixiuyuannuca@126.com

## Abstract

Deep learning-based underwater object detection (UOD) remains a major challenge due to the degraded visibility and difficulty to obtain sufficient underwater object images captured from various perspectives for training. To address these issues, this paper presents a high-precision UOD based on joint self-supervised deblurring and improved spatial transformer network. A self-supervised deblurring subnetwork is introduced into the designed multi-task learning aided object detection architecture to force the shared feature extraction module to output clean features for detection subnetwork. Aiming at alleviating the limitation of insufficient photos from different perspectives, an improved spatial transformer network is designed based on perspective transformation, adaptively enriching image features within the network. The experimental results show that the proposed UOD approach achieved 47.9 mAP in URPC2017 and 70.3 mAP in URPC2018, outperforming many state-of-the-art UOD methods and indicating the designed method is more suitable for UOD.

## 1 Introduction

In recent years, underwater object detection has become more and more important for underwater applications such as ocean exploring and autonomous underwater vehicles [1-3]. However, UOD is still a challenging task because of complicated underwater environments and lighting conditions [4-5]. On one hand, underwater images acquired in complex underwater environments suffer from serious distortion which dramatically degrades the image visibility, affecting the detection accuracy of UOD tasks [6]. On the other hand, in underwater environments, obtaining image data is difficult because it can be accomplished only by trained divers using special equipment. Further, the divers are limited by time and temperature. Thus, it is difficult to obtain sufficient underwater object images captured from various perspectives, limiting the accuracy of deep learning based object detection methods in underwater scenes, where the viewing angle often changes greatly [7].

To improve object detection performance in underwater environments, most of the existing strategies consider underwater image enhancement and UOD tasks as two separate pipelines, where enhancing the visibility of underwater images has been used as a preprocessing step [8]. This separate architecture can improve the accuracy of object detection methods based on shallow-learning classifiers to some extent [9-10], whereas using images preprocessed by deblurring as an input of the deep-learning object detectors does not always guarantee obviously improved object detection performance [11-13]. To ensure image enhancement can improve the accuracy of UOD, some researchers embed an image preprocessing algorithm into the detection network framework [14,5,8], conducting the end-to-end training and recognition in a unified deep learning framework. Fan et al. introduced feature enhancement and anchor refinement into the UOD network, alleviating the adverse effects the blur and texture distortion in underwater environments to a certain extent [14]. Chia-Hung et al. presented a jointly learning image color conversion, which transforms color images to the corresponding grayscale images to solve the problem of underwater color absorption to enhance the object detection performance [5]. Zhang et al. introduced an underwater image enhancement module composed of denoising, color correction and deblurring sub-modules to greatly improve the accuracy of UOD [8]. However, paired training sets are necessary for the training of image enhancement module, leading to difficulty in application in real-world underwater scences. A common problem of these above existing end-to-end methods is that the complex image or feature preprocessing module need to be run when the UOD is conducted, greatly increasing the computational complexity.

To solve the problem of insufficient image samples captured from various complex underwater environments, researchers have explored many methods, which mainly include data augmentation and underwater image synthesis. Data augmentation is a direct and effective way, which generates more training data via making changes to the labeled dataset [15-17]. There are various approaches to augment data, such as adding noise, flipping, cropping, color casting, and blur. These frequently used methods just do simple transformations, which may not be consistent with the special underwater imaging environments, limiting the boosting to UOD. Underwater image synthesis methods can produce a large number of synthetic photos based on physical models or scene prior, simulating a diverse set of water types and degradation levels [18-20]. However, due to the domain gap between synthetic and real data, the deep learning based models trained on synthetic images tend to have a significant performance drop when applied to the real domain. In addition, transfer learning is also an effective way to solve the shortage of underwater images, but it also has the problem of domain gap [21,22]. Although many domain adaptation methods have been proposed, the gap between source and target domain still exists, limiting the accuracy of transfer learning based object detection, and thus transfer learning is always used to initialize the weights of neural networks. In recent years, few-shot object detection has attracted extensive attention as a new way to solve the problem of insufficient samples [23,24]. Most of deep networks for few-shot object detection can effectively decrease the number of samples required, but it is still difficult for these methods to learn the features of objects in a various complex underwater environments because the special conditions of underwater imaging are not considered in the design of these networks.

Aiming at solving the above problems restricting the accuracy of UOD, a high-precision underwater object detection based on joint self-supervised deblurring and improved spatial transformer network (STN) is proposed in this paper according to the characteristics of underwater imaging. Faster R-CNN is employed as the backbone network (denoted as detection subnetwork) and a clean image reconstruction module is integrated into this base architecture to construct visibility enhancement subnetwork, which shares the feature extraction module with the detection subnetwork to enjoy the benefits of multi-task learning. Following the joint optimization scheme, clean features generated by the image debluring subnetwork from blured input images for visibility enhancement can be shared to learn better object classification and object localization in the detection subnetwork, thus boosting the object detection performance in hazy underwater environments. To solve the dependence of traditional supervised learning on real underwater clear images, a transmission map prediction module is designed so that the self-supervised loss for visibility enhancement can be generated based on the model of the underwater imaging.

In order to improve the accuracy of UOD in the case that there are insufficient object images captured from various viewing angles, an improved STN is proposed based on perspective transformation to achieve image feature enrichment adaptively, dealing with the image degradation resulted from different views of camera shooting. The differential improved STN are integrated into the network architecture so that the parameters of image feature adjustment can be learned automatically in an end-to-end fashion, ensuring that the learned object detection network can adapt to various perspectives, with underwater images captured from limited viewing angles.

The experimental results demonstrate that the proposed deep-learning UOD can obtain much more accurate detection results compared with previous works in the literature. The self-supervised image deblurring branch can significantly promote the accuracy of object detection by shared features in the multi-task learning architecture. The presented image feature enrichment module inspired by perspective transformation can obviously improve the performance of the object detection network in various viewing angles.

## 2 Related Works

### A．Underwater object detection

UOD techniques have been employed in marine ecology studies for many years. Strachan et al. [25] utilized color and shape descriptors to detect fish transported on a conveyor belt, which is monitored by a digital camera. Spampinato et al. [26] presented a vision system for detecting fish in real-time videos, which consist of object detection and tracking process. Ravanbakhsh et al. [27] used the Histogram of Oriented Gridients (HOG)+Support Vector Machine (SVM) method to conduct the detection of coral reef fishes. However, the above-mentioned approaches heavily rely on hand-crafted features, which have limited the representation ability.

With the rapid development of deep learning techniques, some deep learning-based UOD techniques [28-37] have also been presented recently. A fast R-CNN-based approach was presented in [30] to conduct the detection and recognition of fish species from underwater images. Moreover, the deep residual network model was adopted in [32] for classifying underwater images of plankton objects. In addition, a lightweight deep neural network was presented in [33] for fish detection via using concatenated ReLU, inception, and HyperNet. Furthermore, a single-shot feature aggregation deep network for UOD was proposed in [35] via introducing multiscale features and complementary context information. The YOLOv2 and YOLOv3 deep models were fine-tuned and tested on the Brackish data set [37], which contains annotated image sequences of fish, crabs, and starfish captured in brackish water with varying visibility. In addition, the datasets for UOD are extremely scarce that hinders the development of deep learning-based UOD techniques.

B. Methods to solve the problem of insufficient underwater images

Adequate training data is necessary for the training of deep learning-based UOD models. Unfortunately, it is difficult to obtain sufficient images data in underwater environments. Many methods have been studied to solve these problems, mainly including data augmentation and image synthesis. Data augmentation increases the amount of data by adding some changes to the labeled dataset. There are various ways to augment data, including adding noise or converting the shapes of images [15-17]. Recent studies have applied data augmentation methods using a generative adversarial neural network to domain adaptation [38].

Image synthesis methods are proposed to synthesize distorted underwater images from high quality in-air RGB or RGB-D images. Image synthesis approaches can be broadly classified into two categories: physical model-based and deep learning-based methods. Physical model-based methods synthesize underwater images via using an underwater image formation model [18, 39, 40]. Recently, Generative Adversarial Networks [41-44] have been investigated in the underwater image synthesis field due to its successes in image-to-image translation tasks. Li et al [44] treated underwater image synthesis as an image-to-image translation task and exploited a single GAN to synthesize underwater images from in-air RGB-D images. To alleviate the needs for training image pairs, Fabbri et al. [42] applied a two-way Cycle-Consistent Adversarial Networks, which allows learning the mutual translation between in-air and underwater domains from unpaired images. However, both physical model-based and deep learning-based underwater image synthesis methods cannot accurately model the degradation progress of underwater imaging, resulting in unsatisfactory synthetic images. The commonly used physical underwater image formation model can only synthesize 10 Jerlov water types and considers only two factors in the degradation progress, leading to significant errors in the generated images. The capability of GAN-based methods is also limited to model the underwater environments. Moreover, Generative Adversarial Networks always encounter the model collapse problem, generating images with monotonous colors and frequent artifacts.

## 3 Proposed methods

### A. The overview of the proposed framework

As shown in Fig. 1, the proposed network architecture consists of two subnetworks: a detection subnet and an image processing subnet, which includes two branches: a transmission map prediction branch and an image debluring branch. The detection subnet adopts Faster R-CNN as the base architecture, which shares the feature extraction module with the image debluring branch and is responsible for object classification and object localization. The image debluring branch is designed by attaching a clean image reconstruction module to the feature extraction module for visibility enhancement. The detection subnet and image debluring branch share the feature extraction module to ensure that the clean features produced can be used in both subnetworks during joint learning. The self-supervised loss for image debluring is generated based on the predicted transmission map and the calculated background light, according to the model of the underwater imaging. The proposed image feature enrichment module is integrated into the detection subnet to achieve the adaptive perspective transformation of image features, promoting the accuracy of UOD in various viewing angles. The whole network can be trained end-to-end and objects can be predicted by using the detection subnet, maintaining fast prediction running time.

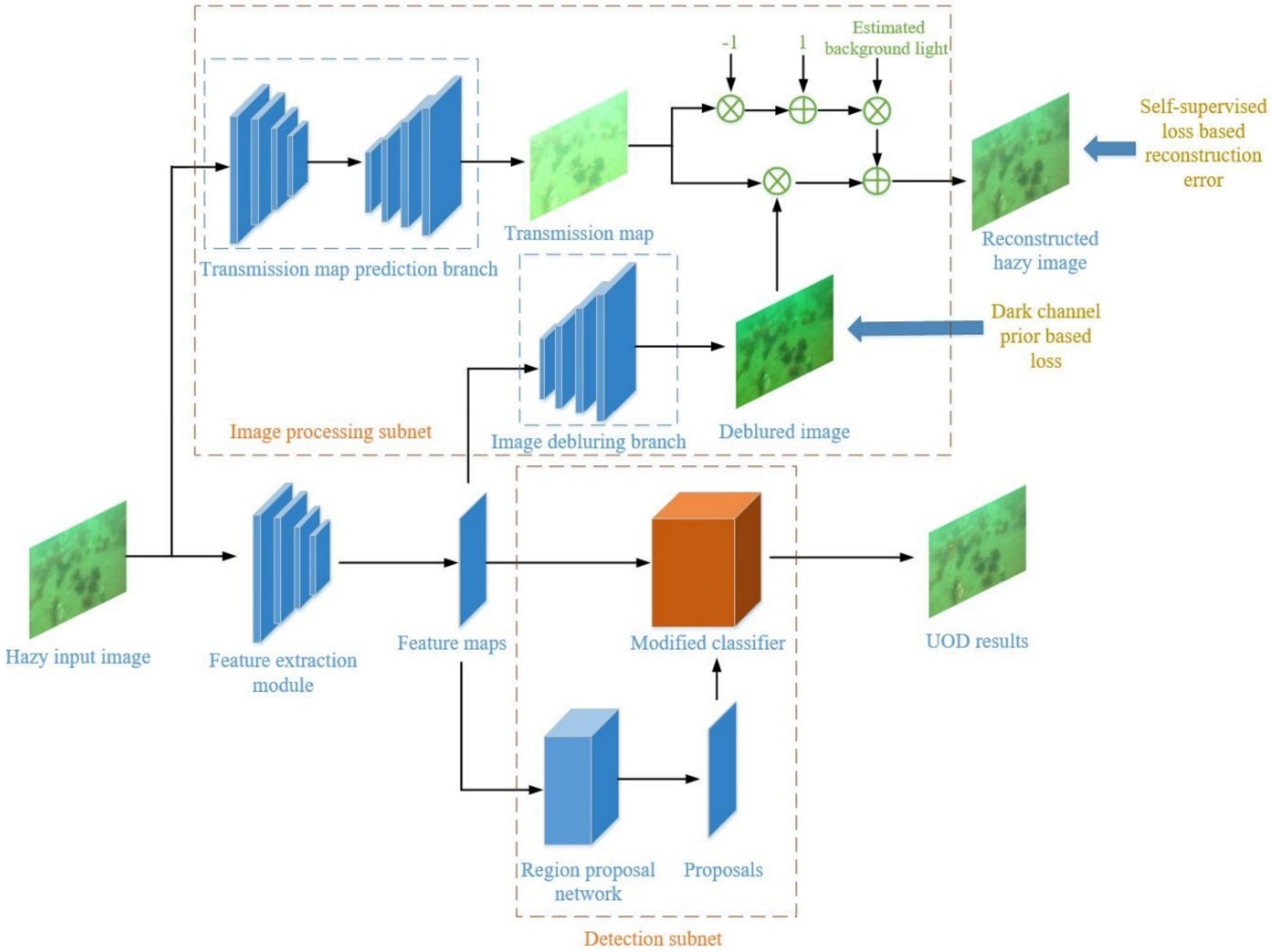

Figure 1. The framework of the proposed UOD methods in this paper

B. Principle of self-supervised image debluring

In 1980, Mc Glamery proposed a classical model of the underwater imaging system in [45], which pointed out that water and a large number of suspended particles can absorb and scatter light, and make underwater images possess colour distortion, blurred detail, low overall brightness, and low contrast. According to the underwater optical imaging model, the underwater image formation can be expressed as follow:

$$I(x) = J(x)t(x) + A(1 - t(x)) \tag{1}$$

where $x$ is a pixel, $I$ is the blured underwater image, $J$ is the clear image, $A$ is the background light, and $t(x)=e^{-\beta d(x)}$ is the underwater transmission map which represents light attenuation due to the scattering medium. The level of attenuation is determined by the attenuation coefficient $\beta$ and scene depth $d(x)$.

Inspired by the image formation model in Eq. (1), the hazy image can be reconstructed based on the predicted transmission map, clear image and background light. Thus, the self-supervised loss for image debluring is generated from the differences between the original and reconstructed hazy images, and can be expressed as:

$$L_{\mathrm{Rec}} = \left\| I_{\mathrm{Rec}}(x) - I_{ori}(x) \right\| \tag{2}$$

where $\|\bullet\|$, $I_{ori}(x)$ and $I_{\mathrm{Rec}}(x)$ denote Frobenius norm of a given matrix, original image and reconstructed image, respectively.

The reconstruction loss can constrain the reconstructed blured images to be well close to the real blured images, however, cannot ensure that the predicted clear images are always consistent with the real clear images. Therefore, a loss based on dark channel prior (DCP) is introduced in this paper to further constrain the clear image prediction branch. DCP is a statistical conclusion, which points out that in patches on an outdoor haze-free image, there is always one channel at least having some pixels whose intensity are very low and close to zero [46]. Due to the similarity of imaging underwater and in the air, the basic conclusion of DCP can be also applied to image debluring in underwater environments.

The transmission map prediction branch consists of a VGG19 feature extractor, 4 up convolution layers and a final convolution layer which outputs the predicted transmission maps at original input resolution. As the background light is independent of the image content and owns the global property, it is estimated directly in this paper via the robust statistical model [46], thus reducing the training difficulty of the whole network.

Through incorporating $t(x)$ and $A$ into one variable, a transformed formula can be gained from the underwater image model in Eq. (1) as:

$$J(x) = B(x)I(x) - B(x) + 1 \tag{3}$$

where $B(x) = \dfrac{\dfrac{1}{t(x)}(I(x) - A) + (A - 1)}{I(x) - 1}$

Inspired by [47], the formula in Eq. (3) is adopted as the basis of the design for the image debluring branch, in which $B(x)$ is estimated firstly and $J(x)$ is predicted sequentially. The clean image reconstruction module of the image debluring branch consists of three submodules: upsampling submodule, multiscale mapping submodule and image generation submodule, as can be seen in Fig. 2.

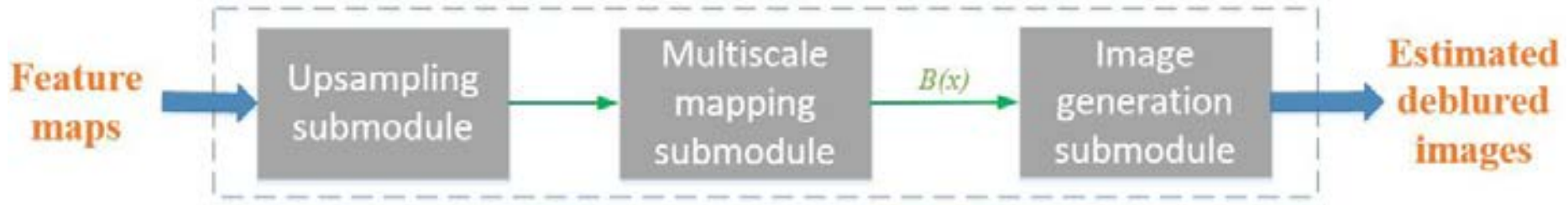

Figure 2. The architecture of the clean image reconstruction module

In [48], the bilinear interpolation technique has been proven to be effective in the CNN-based blur removal approach, where the pooled feature maps are bilinearly upsampled to prepare for generating the dehazed output. Inspired by this method, the upsampling submodule first employs a $1 \times 1$ convolutional layer to reduce the feature dimensions. Afterwards, a bilinear interpolation is used to enlarge the size of the feature maps to the same size as the input images.

In consideration of the good performance of multiscale feature extraction in blur removal methods [49], the multiscale mapping submodule employs convolution layers in parallel based on the Inception architecture [50] to extract multiscale features for enhancing visibility. The multiscale mapping submodule is composed of four parallel convolutions, namely, one $1 \times 1$ convolution, one $3 \times 3$ convolution, one $5 \times 5$ convolution, and one $7 \times 7$ convolution with the same number of channels, which is set at 4. The resulting features are fused and followed by another $3 \times 3$ convolution to estimate $B(x)$.

The image generation submodule takes $B(x)$ as input and adopts an element-wise multiplication layer, an elementwise subtraction layer, and an element-wise addition layer to calculate the transformed formula in Eq. (3), thus obtaining the predicted clear underwater images.

The features of the input images extracted by the feature extraction module can be degraded by blur, thereby leading to poor object detection performance. The designed image debluring branch is embedded into the base architecture to generate clean shared features from the feature extraction module via learning the visibility enhancement task. Consequently, the detection performance is improved by jointly optimizing visibility enhancement, object classification, and object localization.

C.Principle of the improved STN and image feature enrichment

Different shooting angles of underwater objects show different visual effects, however, it is difficult and two costly to obtain a large number of photos of objects shooted from various angles. The basic spatial transformer network (STN) proposed by [51] can conduct an affine transformation of images in the network so that the trained network has a certain degree of spatial transformation invariance without collecting a large number of samples from a variety of different angles. Affine transformation is a linear transformation between 2-dimensional coordinates and can only implement translation, scaling, flipping, rotation, and shearing of the pictures. However, the transformation between different shooting angles is a transformation between 3-dimensional coordinates, implying that the basic STN based on affine transformation possess limited improvement in the accuracy of UOD in various viewing angles.

Unlike affine transformation, perspective transformation can be treated as 3D projection, which just like viewing objects through a camera viewfinder. The performance of the perspective transformation can be adjusted well according to the camera's position, orientation, and field of view, implying that perspective transformation has a better visual effect in simulating different shooting angles of images than affine transformation. Thus, different from the basic STN, an improved

STN based on perspective transformation is proposed in this paper to improve the detection accuracy of underwater objects from different angles of view with a limited number of samples in the corresponding environments.

The basic STN is composed of three parts: a localization network, a grid generator and a sampler, which implement successively parameter prediction, coordinate mapping, and pixel acquisition, respectively. The localization network takes the input feature maps and outputs the parameters of the spatial transformation via a number of hidden layers. Then, the predicted transformation parameters are used in the grid generator to create a sampling grid, which is a set of points where the input feature maps should be sampled to produce the transformed outputs. Finally, the feature maps and the sampling grid are taken as inputs to the sampler, producing the output feature maps sampled from the input at the grid points.

The localization network of the improved STN in this paper is designed to predict the eight transformation parameters, which are used in the following modified grid generator. The pointwise transformation in the modified grid generator of the proposed STN is conducted based on Eq. (4) to complete the mapping based on perspective transformation. The sampler in this paper achieves the differentiable image sampling via a bilinear sampling kernel just like the classical STN.

$$\begin{pmatrix} x \\ y \\ z \end{pmatrix} = \begin{pmatrix} \theta_{11} & \theta_{12} & \theta_{13} \\ \theta_{21} & \theta_{22} & \theta_{23} \\ \theta_{31} & \theta_{32} & \theta_{33} \end{pmatrix} \begin{pmatrix} x_i^t \\ y_i^t \\ 1 \end{pmatrix}$$
$$x_i^s = \frac{x}{z} \qquad (4)$$
$$y_i^s = \frac{y}{z}$$

where $\left(x_i^t, y_i^t\right)$ are the target coordinates of the regular grid in the output feature map, $\left(x_i^s, y_i^s\right)$ are the source coordinates in the input feature map that define the sample points, and $\left(x, y, z\right)$ is the intermediate variable of perspective transformation.

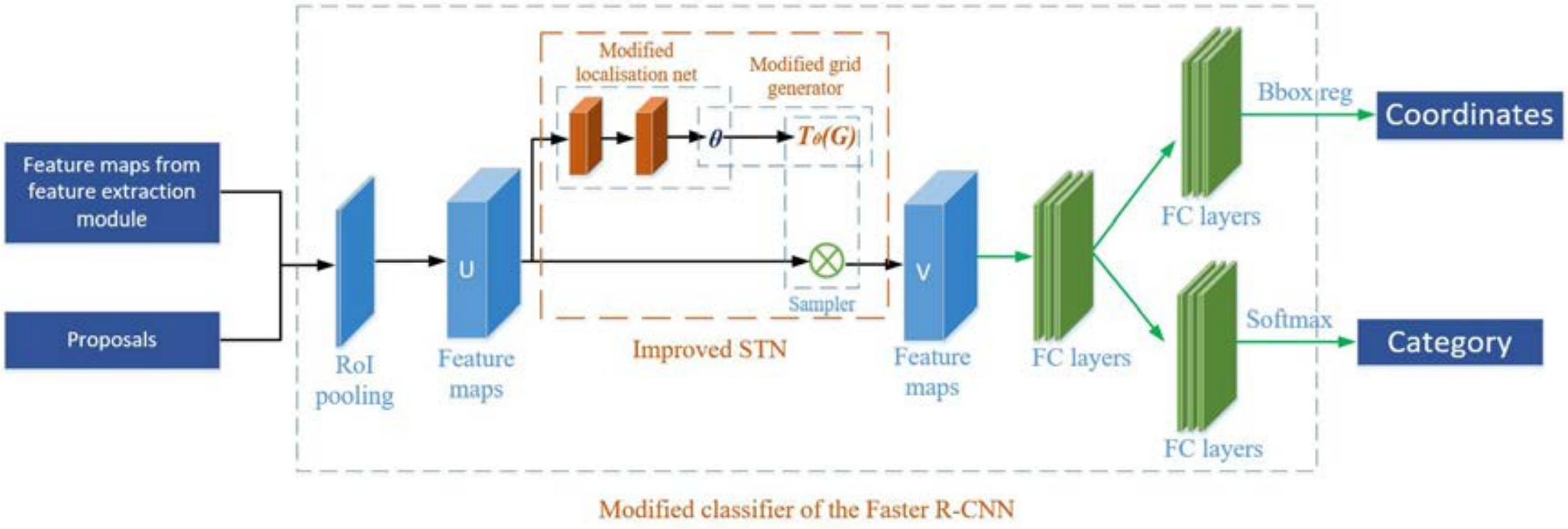

Figure 3. The principle of image feature enrichment via the improved STN

As shown in Fig.3, the image feature enrichment is conducted via the improved STN, which is inserted into the modified classifier of the Faster R-CNN. The improved STN takes the feature maps $U$ after RoI pooling as the input, and then the modified localization network of the STN estimates the deformation parameters, which are in succession used as the input of the mesh generator and sampler. The outputs of the STN are the deformed feature maps $V$, which are then propagated forward to the higher-level layers of Faster R-CNN to calculate the softmax loss. The transformation mode of the space transformer depends on each sample, and each input sample can generate a suitable spatial transformation for its image or feature maps, so that the Faster R-CNN is more robust to various forms of perspective transformation in underwater environments.

## 4 Experimental Results and Discussions

### A. Datasets

The proposed method was evaluated on the publicly-available two underwater datasets Underwater Robotic Picking Contest 2017 (URPC2017) and Underwater Robotic Picking Contest 2018 (URPC2018), collected on the natural seabed at Zhangzidao, Dalian, China. The URPC2017 dataset has 18,982 training images and 983 testing images, including 3 object categories: seacucumber, seaurchin and scallop. URPC2018 is composed of 2,901 aquatic images for training and 800 samples for testing, containing 4 categories: holothurian, echinus, scallop and starfish. Both the two datasets provide underwater images and bounding box annotations.

### B. Training details

All the networks in the ablation studies were trained with the Adam optimization algorithm on a single NVIDIA 3090 Ti GPU with a 24 GB memory. The scale of input images for both training and testing was set as $512\times512$. During the training of both the two datasets, dropout of 0.3 was applied to avoid over-fitting. For URPC2017, the batch-size is 16, the learning rate is 0.0001, and all the networks obtain the best performance after running 150 epochs. For URPC2018, the batch-size is 16, and all the networks achieve convergence after running 100 epochs with a high learning rate 0.001.

### C. Design of ablation and comparative experiments

In order to verify the improvement effects of the proposed joint self-supervised deblurring branch and integrated image feature enrichment module inspired by perspective transformation, some variations of the proposed deep-learning underwater object detection were designed via removing the corresponding parts in the ablation study. The different methods and variations designed for the ablation study are listed as below:

Baseline: the classical Faster R-CNN;

Faster R-CNN deblurring: Baseline only with joint self-supervised deblurring branch;

Faster R-CNN classical STN: Baseline only with classical STN;

Faster R-CNN improved STN: Baseline only with improved STN based on perspective transformation;

Proposed complete method: Baseline with joint self-supervised deblurring branch and improved STN based on perspective transformation.

Besides, some representative existing strategies, considering visibility enhancement and UOD as two separate pipelines, were tested on the two datasets, to further validate the performance of the proposed method which deals with visibility enhancement and UOD in a unified architecture. Three outstanding deblurring methods for underwater images, namely, MSDB [52], GAN-RS [53], and UWCNN [54] were selected to combine with the base detection network (Faster R-CNN) for forming three combination models called MSDB + Faster R-CNN, GAN-RS + Faster R-CNN, and UWCNN + Faster R-CNN, respectively.

D. Effects of joint self-supervised deblurring branch

The quantitative and qualitative results of ablation experiments are shown in Tab.1 and Fig.4, respectively. As illustrated in Tab.1, the accuracy rates obtained by using the Faster R-CNN deblurring in URPC2017 are 33.6 AP for the seacucumbers, 50.7 AP for the seaurchins and 36.9 AP for the scallops, which are 6.8, 2.8 and 4 higher than those provided by the Baseline. For URPC2018, the AP of the Faster R-CNN deblurring are 50.1 AP for the holothurians, 85.9 AP for the echinus, 40.6 AP for the scallops and 81.3 AP for the starfishes, which are 5.3, 3.2, 5.2 and 3.1 higher than that of the Baseline. Both of the experimental results in URPC2017 and URPC2018 indicate that the joint self-supervised deblurring branch can obviously improve the accuracy of UOD without the ground truth for underwater image visibility enhancement. The self-supervised loss generated by the imaging model and DCP can well accomplish the training of the deblurring branch, further improving the accuracy of UOD via the shared feature extraction module. The main reason behind such performance improvement without additional computation is that: during the training process, in order to let clean image reconstruction module produce a clear blur-free image from the shared features, the feature extraction module will learn to extract features that contain more blur-free spatial-related information to minimize the reconstruction loss. While the shared features contain more blur-free spatial information, the detection subnetwork is able to use these features to accurately detect objects in blurred images.

Table 1. The quantitative results of ablation experiments

(a) The quantitative results in URPC2017

| Method | Seacucumber | Seaurchin | Scallop | mAP |
|---|---|---|---|---|
| Baseline | 26.8 | 47.9 | 32.9 | 35.9 |
| Faster R-CNN deblurring | 33.6 | 50.7 | 36.9 | 40.4 |
| Faster R-CNN classical STN | 32.4 | 50.1 | 36.1 | 39.5 |
| Faster R-CNN improved STN | 38.3 | 53.5 | 40.4 | 44.1 |
| Proposed complete method | 44.6 | 55.7 | 43.5 | 47.9 |

(b) The quantitative results in URPC2018

| Method | Holothurian | Echinus | Scallop | Starfish | mAP |
|---|---|---|---|---|---|
| Baseline | 44.8 | 82.7 | 35.4 | 78.2 | 60.3 |
| Faster R-CNN deblurring | 50.1 | 85.9 | 40.6 | 81.3 | 64.5 |
| Faster R-CNN classical STN | 48.4 | 85.1 | 38.6 | 80.5 | 63.2 |
| Faster R-CNN improved STN | 52.9 | 88.5 | 43.2 | 83.6 | 67.1 |
| Proposed complete method | 57.6 | 90.5 | 47.4 | 85.8 | 70.3 |

E. Effects of the designed image feature enrichment module

Faster R-CNN classical STN can obtain 3.6 for URPC2017 and 2.9 for URPC2018 higher mAP than the Baseline, implying that the classical STN based image feature enrichment module can promote the robustness of UOD in various shooting angles to some extent. Compared with Faster R-CNN classical STN, Faster R-CNN improved STN improves the UOD results by 4.6 for URPC2017 and 3.9 for URPC2018 higher mAP, indicating that the improved STN based on perspective transformation can further facilitate the robustness of the detector to the change of object shape caused by different shooting angles through adaptive training.

In fact, the difference of the shooting angles between the training images and testing images in the two datasets is not so large that the improved STN can fully demonstrate its superiority in image feature enriching. Provided that the difference between the training and testing data is bigger than that of the two adopted datasets, the advantage of the proposed improved STN may be more obvious with a larger scale of detection accuracy improvement. The improved STN based on perspective transformation can help the detector to adaptively learn the features of objects in the environments where the shooting angles are different from that of training data to a certain extent.

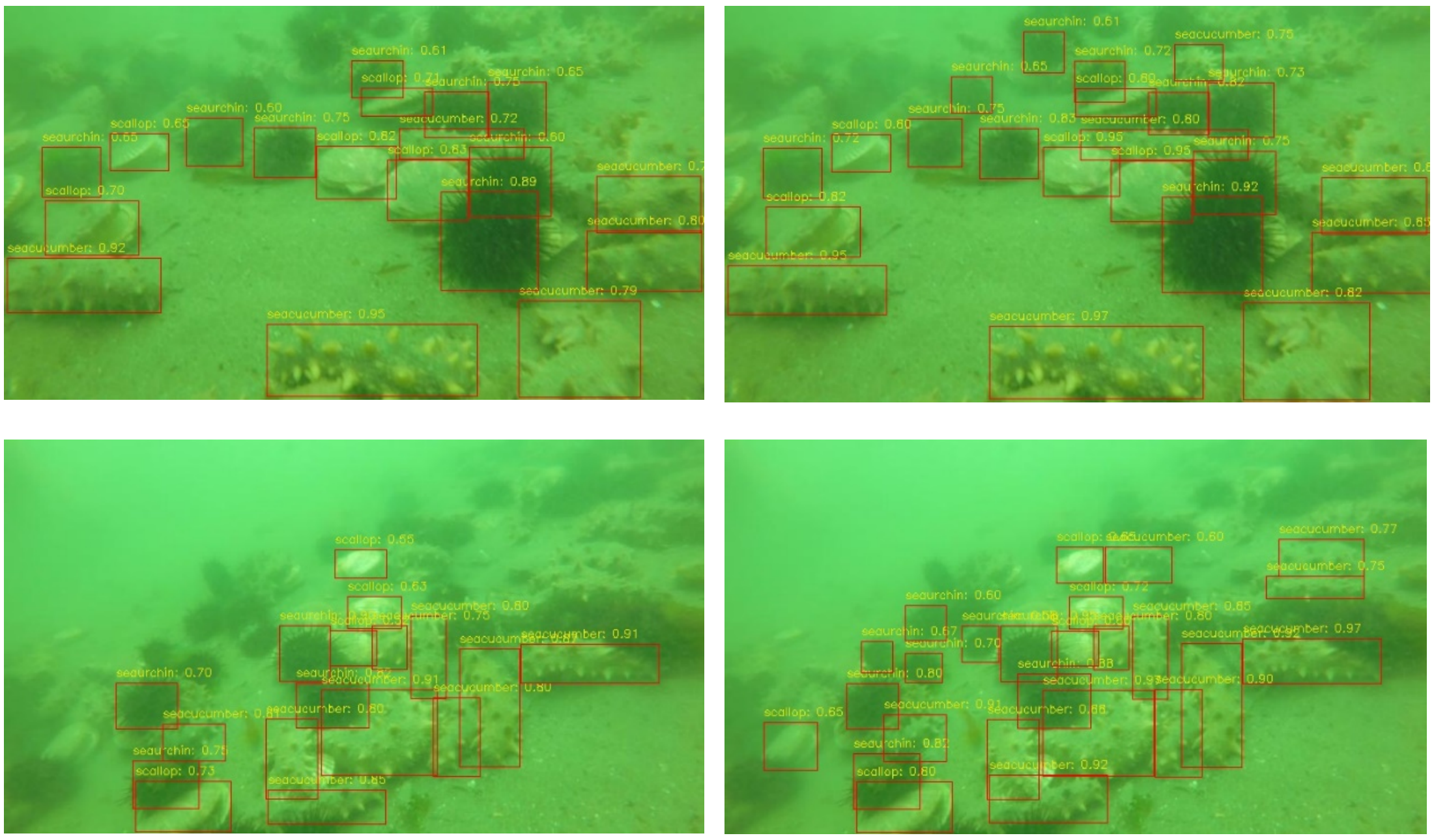

(a) The qualitative examples of URPC2017

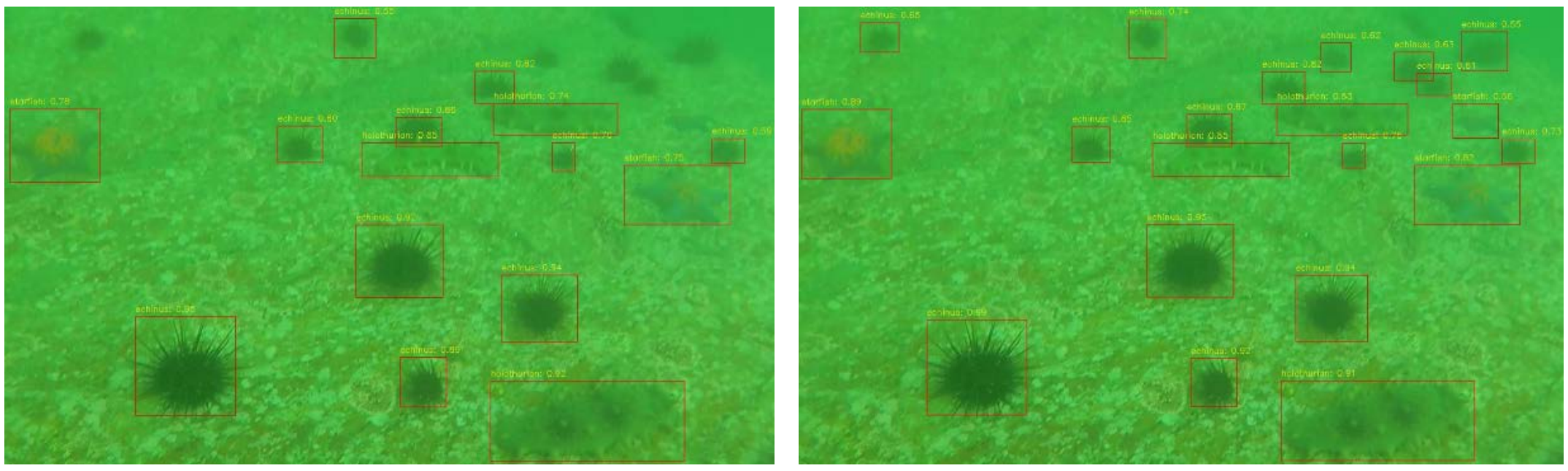

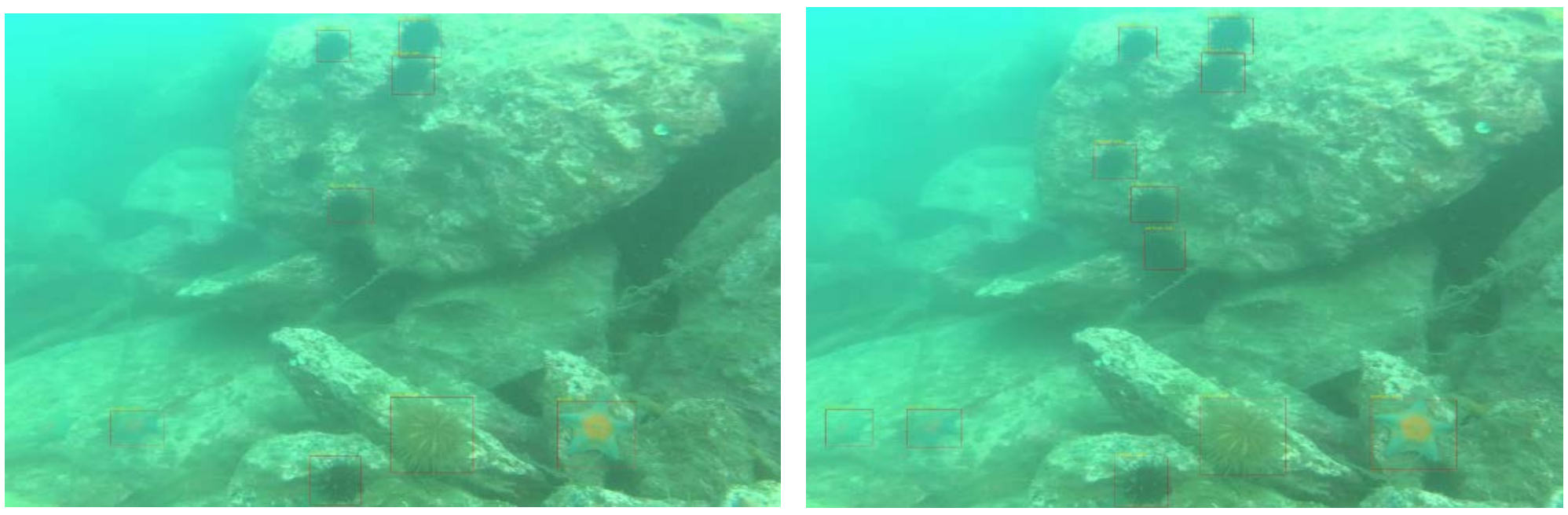

(b) The qualitative examples of URPC2018

Figure 4. The qualitative examples of the Baseline (left) vs. the Proposed complete method (right)

## F. Comparisons of methods based on separated and associated debluring

The experimental performance of MSDB + Faster R-CNN was obtained through the following consecutive steps. Firstly, all the training samples of URPC2017 and URPC2018 were processed via using the MSDB method to generate the corresponding blur-free training sets. Afterwards, the Faster R-CNN was trained on the images of the corresponding blur-free training sets to produce the models of the detectors for MSDB + Faster R-CNN. Finally, the original testing images were preprocessed to achieve visibility enhancement via the MSDB method, and then the enhanced images were inputed into the trained models of the detectors to acquire the UOD results of MSDB + Faster R-CNN. The performance verification process of GAN-RS + Faster R-CNN and UWCNN + Faster R-CNN is similar to that of MSDB + Faster R-CNN except that the image deblurring method MSDB is replaced by GAN-RS and UWCNN, respectively. The quantitative and qualitative results of comparative experiments are shown in Tab.2 and Fig.5, respectively.

Table 2. The quantitative results of comparative experiments

(a) The quantitative results in URPC2017

| Method | Seacucumber | Seaurchin | Scallop | mAP |
|---|---|---|---|---|
| Baseline | 26.8 | 47.9 | 32.9 | 35.9 |
| Faster R-CNN deblurring | 33.6 | 50.7 | 36.9 | 40.4 |
| MSDB + Faster R-CNN | 28.5 | 48.8 | 34.1 | 37.1 |
| GAN-RS + Faster R-CNN | 29.9 | 49.7 | 35.2 | 38.3 |
| UWCNN + Faster R-CNN | 25.7 | 46.9 | 32.1 | 34.9 |
| Proposed complete method | 44.6 | 55.7 | 43.5 | 47.9 |

(b) The quantitative results in URPC2018

| Method | Holothurian | Echinus | Scallop | Starfish | mAP |
|---|---|---|---|---|---|
| Baseline | 44.8 | 82.7 | 35.4 | 78.2 | 60.3 |
| Faster R-CNN deblurring | 50.1 | 85.9 | 40.6 | 81.3 | 64.5 |
| MSDB + Faster R-CNN | 47.1 | 83.7 | 37.5 | 79.1 | 61.9 |
| GAN-RS + Faster R-CNN | 48.3 | 84.1 | 37.9 | 79.7 | 62.5 |
| UWCNN + Faster R-CNN | 41.5 | 80.4 | 31.7 | 76.8 | 57.6 |
| Proposed complete method | 57.6 | 90.5 | 47.4 | 85.8 | 70.3 |

As shown in Tab.2, the results of the three UOD methods based on separated debluring are close to or slightly worse than that of the baseline which is trained and tested directly on the original hazy datasets, indicating that decoupled image visibility enhancement cannot guarantee to improve the accuracy of subsequent deep learning based object detection. The mAP of the Faster R-CNN deblurring is obviously better than that of the three UOD methods based on separated debluring, implying that coupled visibility enhancement is assuredly effective to promote the accuracy of deep learning based object detection. Above all, the proposed complete method in this paper can obtain the best results as illustrated in Tab.2, showing that the designed method is better suited for UOD.

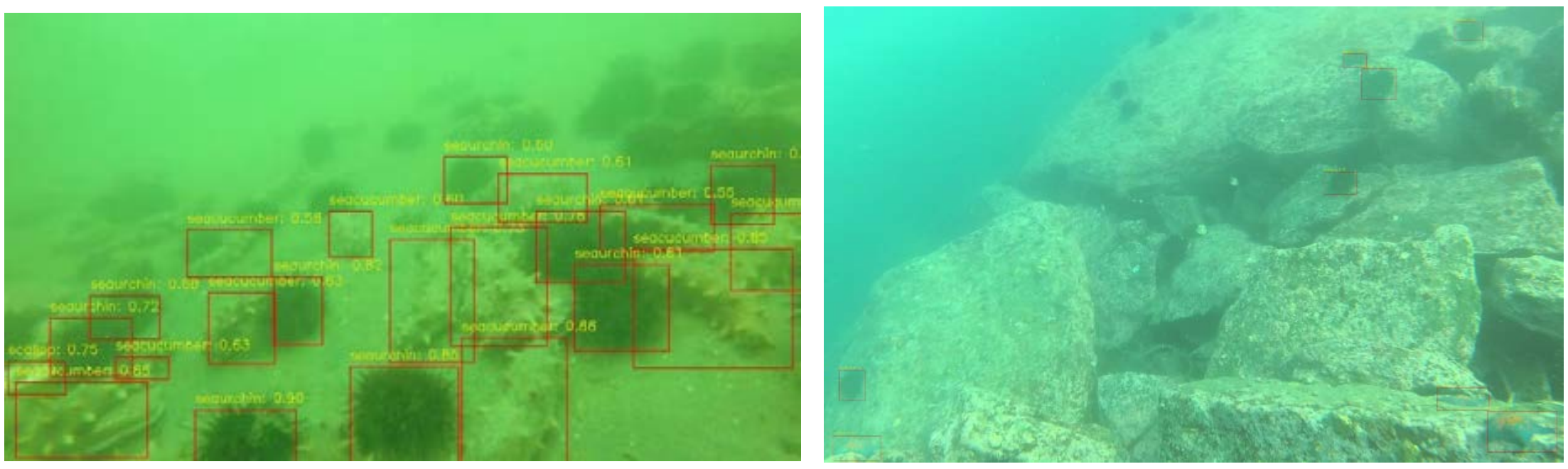

(a) The results via the classical Faster R-CNN

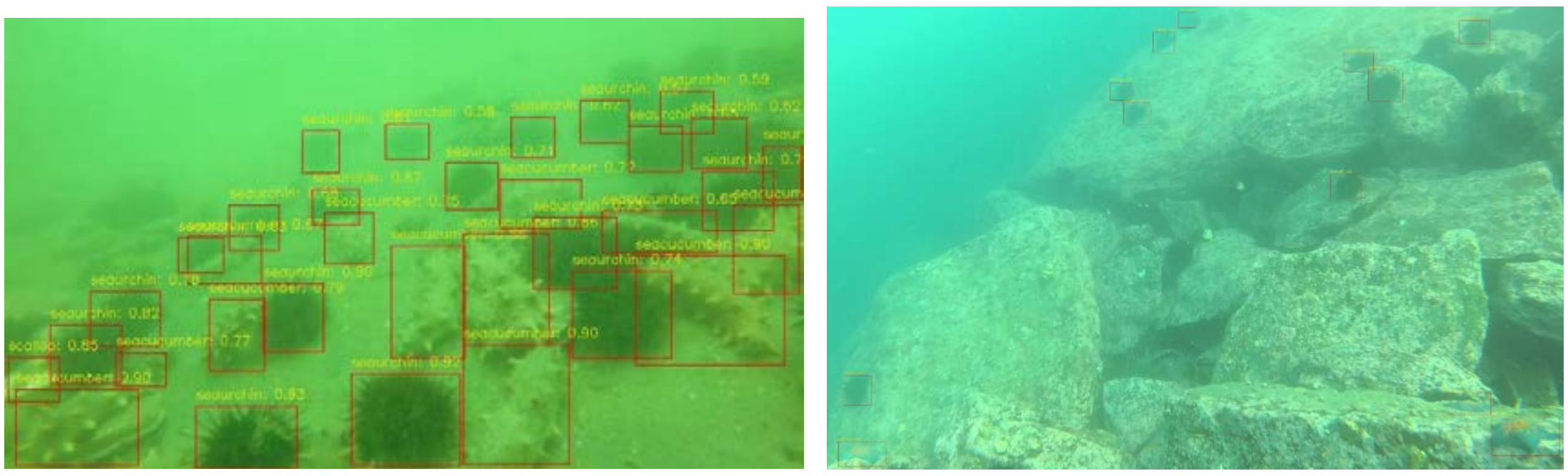

(b) The results via Faster R-CNN deblurring

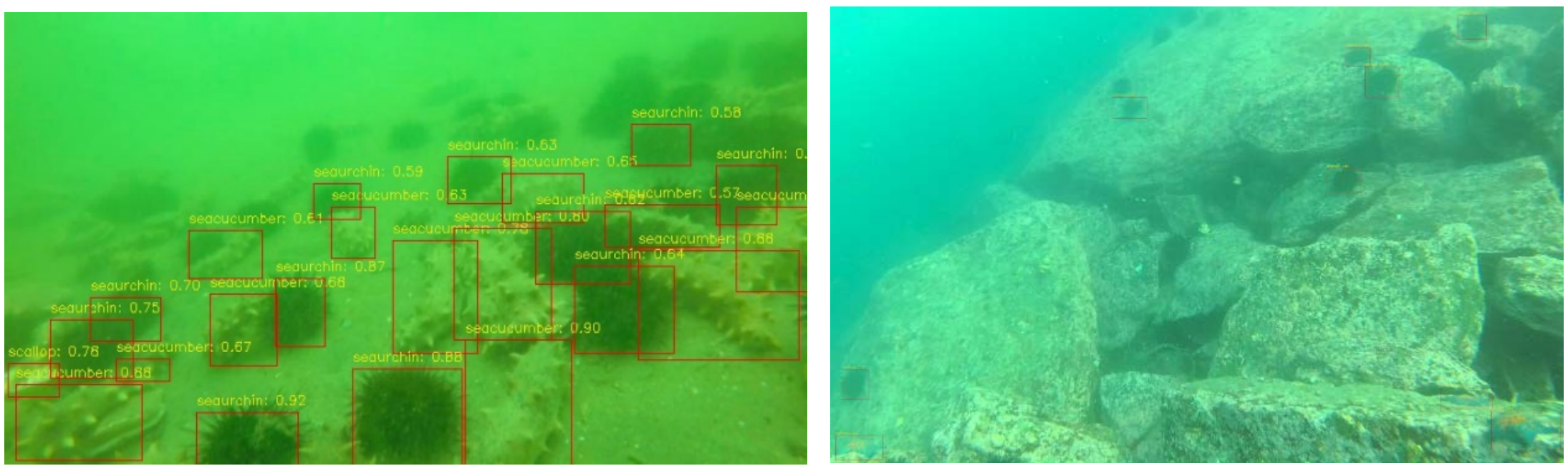

(c) The results via MSDB + Faster R-CNN

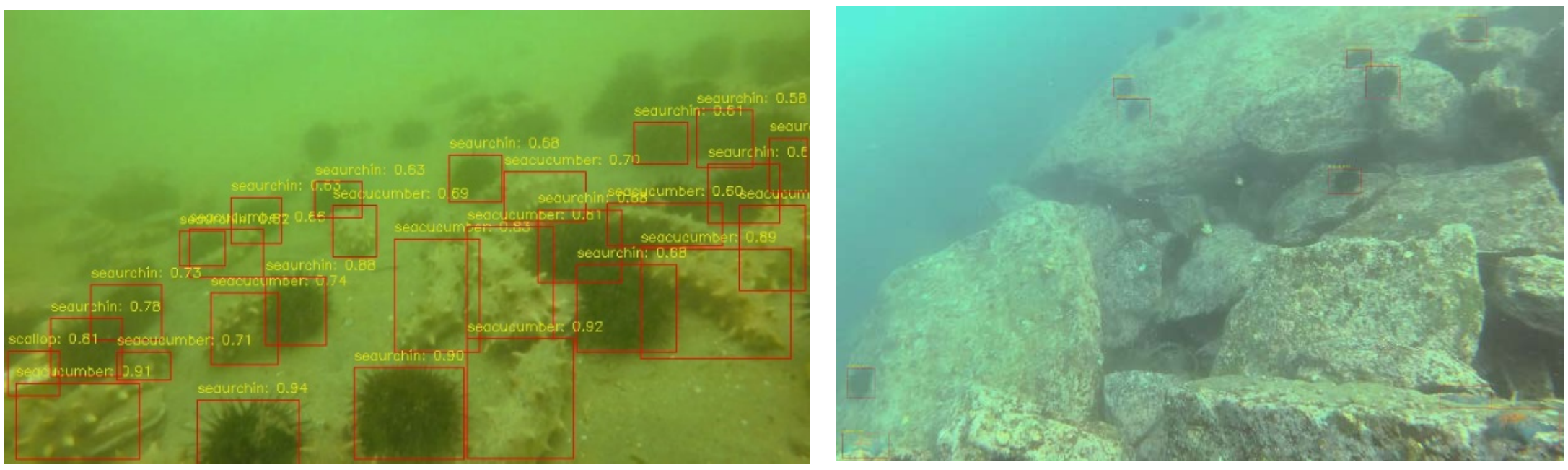

(d) The results via GAN-RS + Faster R-CNN

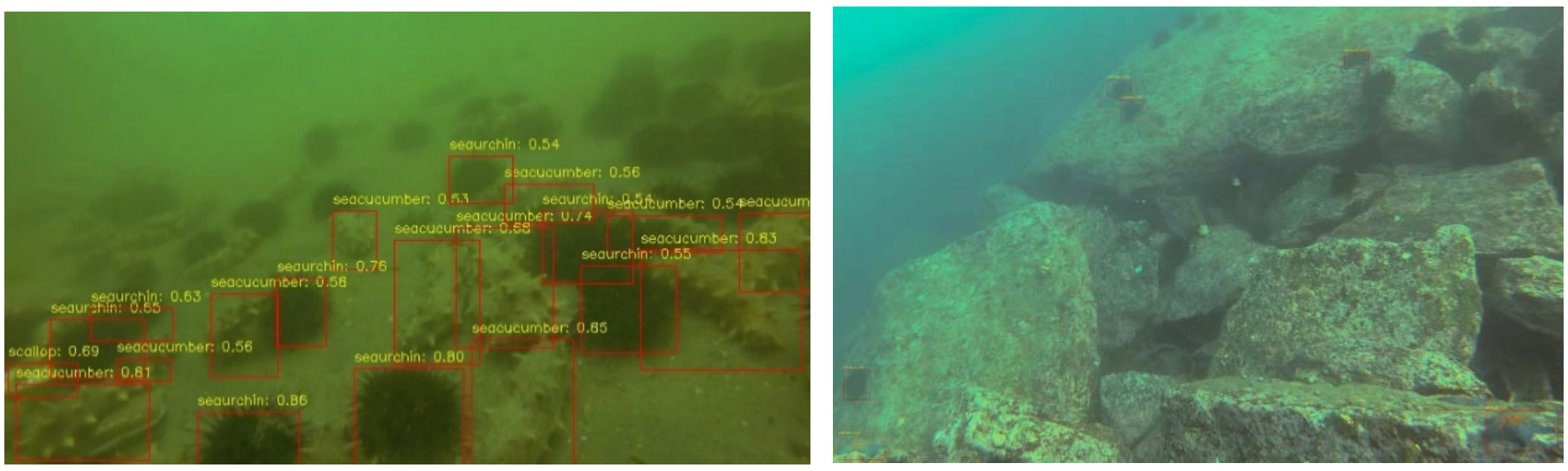

(e) The results via UWCNN + Faster R-CNN

Figure 5. The qualitative examples of comparative experiments. The images in the left and right column are from URPC2017 and URPC2018, respectively. The images in the same column from top to bottom are the results of the classical Faster R-CNN, Faster R-CNN deblurring, MSDB + Faster R-CNN, GAN-RS + Faster R-CNN and UWCNN + Faster R-CNN, respectively.

## 5 Conclusion

In the past few years, deep learning-based object detection methods have been widely studied and successfully applied in many computer vision applications. However, detecting objects in underwater environments remains a major challenge due to the degraded visibility and insufficient underwater object images captured from various perspectives for training.

To address these issues, this paper presents a high-precision UOD based on joint self-supervised deblurring and improved spatial transformer network. A multi-task learning aided object detection architecture is designed, including detection subnetwork and self-supervised deblurring subnetwork, which is introduced to force the shared feature extraction module to output clean features for detection subnetwork. To alleviate the limitation of insufficient photos from different perspectives on the training, an improved spatial transformer network based on perspective transformation is proposed, adaptively enriching image features within the network.

The ablation experiments verified the effectiveness of the designed joint self-supervised deblurring subnetwork and improved spatial transformer network. The experimental results show that the proposed UOD approach achieved 47.9 mAP in URPC2017 and 70.3 mAP in URPC2018, which outperform some representative UOD in the literature and indicate the designed method is more suitable for UOD.

Acknowledgments

This work was supported by the Talent Award Scheme of Shanxi Province (Grant no. 304/18001618) and the Science Foundation of North University of China (Grant no.304/20170024).